\documentclass{article}

\usepackage{PRIMEarxiv}

\usepackage[utf8]{inputenc} 
\usepackage[T1]{fontenc}    
\usepackage{hyperref}       
\usepackage{url}            
\usepackage{booktabs}       
\usepackage{amsfonts}       
\usepackage{nicefrac}       
\usepackage{microtype}      
\usepackage{lipsum}
\usepackage{fancyhdr}       
\usepackage{graphicx}       
\graphicspath{{media/}}     
\usepackage{amsmath}

\title{BG-YOLO: A Bidirectional-Guided Method for Underwater Object Detection
}

\author{
  Jian Zhang$^{1,2+*}$, Ruiteng Zhang$^{3+}$, Xinyue Yan$^{1}$, Xiting Zhuang$^{1}$, Ruicheng Cao$^{4}$\\
  {1} School of Tropical Agriculture and Forestry, Hainan University, Danzhou 571700, China \\
  {2} School of Information and Communication Engineering, Haikou 570100, China\\
  {3} College of Computer Science and Technology, Zhejiang University, Hangzhou 310000, China\\
  {4} School of Cybersecurity, Northwestern Polytechnical University, Xian 710000, China\\
  {+} First two authors contributed equally\\
  \texttt{\{Jian Zhang\}whealther@hainan.edu.cn} 
}

\begin{document}
\maketitle

\begin{abstract}
Degraded underwater images decrease the accuracy of underwater object detection. However, existing methods for underwater image enhancement mainly focus on improving the indicators in visual aspects, which may not benefit the tasks of underwater image detection, and may lead to serious degradation in performance. To alleviate this problem, we proposed a bidirectional-guided method for underwater object detection, referred to as BG-YOLO. In the proposed method, network is organized by constructing an enhancement branch and a detection branch in a parallel way. The enhancement branch consists of a cascade of an image enhancement subnet and an object detection subnet. And the detection branch only consists of a detection subnet. A feature guided module connects the shallow convolution layer of the two branches. When training the enhancement branch, the object detection subnet in the enhancement branch guides the image enhancement subnet to be optimized towards the direction that is most conducive to the detection task. The shallow feature map of the trained enhancement branch will be output to the feature guided module, constraining the optimization of detection branch through consistency loss and prompting detection branch to learn more detailed information of the objects. And hence the detection performance will be refined. During the detection tasks, only detection branch will be reserved so that no additional cost of computation will be introduced. Extensive experiments demonstrate that the proposed method shows significant improvement in performance of the detector in severely degraded underwater scenes while maintaining a remarkable detection speed. 
\end{abstract}


\section{Introduction}
\begin{figure}[ht!]
    \centering
    \includegraphics[width=0.6\linewidth]{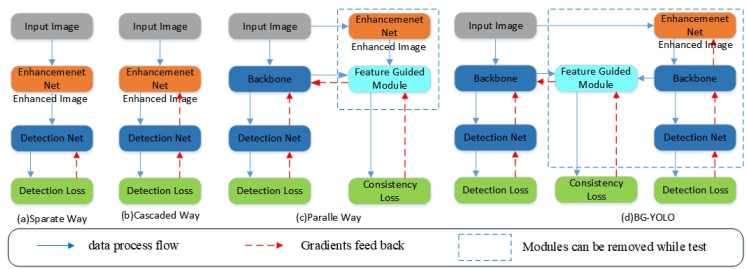}
    \caption{Different combination ways of underwater image enhancement and object detection.}
    \label{fig1:enter-label}
\end{figure}

In underwater scenes, the images suffer from wavelength-related light absorption and scattering, which results in serious degradation, and the accuracy of underwater object detection tasks is affected. Some researches use underwater image enhancement methods to process degraded images, refining the quality of images to improve the accuracy of underwater object detection tasks. However, image enhancement tasks and object detection tasks have different goals and indicators, which leads to the differences in optimization studies and optimal solutions [1]. Therefore, adopting image enhancement as the pre-processing directly may not effectively improve the accuracy of object detection model [2]. Many researches began to focus on the combination of underwater image enhancement network and object detection network to improve the accuracy of underwater object detection. Liu et al. [9] classified the combination of the two into three ways: the separate way, the cascaded way and the parallel way, as is shown in Fig. \ref{fig1:enter-label}(a)-(c).

In the cascaded way (shown in Fig. \ref{fig1:enter-label}(b)) the image enhancement network and the object detection network are integrated into one single pipeline, which ensures that the two individual tasks are optimized in a common direction of both through joint optimization. In DE-YOLO [5] and the document [6], image enhancement module and detection module are integrated into one single framework in cascaded way, and relevant detection information from detector is used to guide the enhancement module to optimize in the direction benefiting the detection tasks, which improve the accuracy of object detection. Organizing the two networks in the cascaded way improves the performance of the detection tasks, however, introduces additional computational cost in test stage.

Different from the two mentioned above, in the parallel way (shown in Fig. \ref{fig1:enter-label}(c)), image enhancement network and object detection network are organized in a parallel way, enhanced images is used to guide the training of the detection network, with the purpose of ameliorating the performance of object detection in various degraded scenes [7-8]. Liu et al. [9] organized enhancement branch and detection branch in a parallel way, and introduced a feature guided module guiding the shallow layers of detection branch to learn the lost details of objects with enhanced images. During the test stage, the enhancement branch and feature guided module will be removed hence no additional cost of computation will be introduced. Comparing to the cascaded way, in the parallel way only the detection branch will be reserved so that no additional cost will be introduced. However, Liu et al. adopted individually trained enhancement network, which cannot ensure that the images used for guided are always conducive to the object detection tasks.

In response to the limitations mentioned above, we attempt to combine the advantages of the cascaded and the parallel way, and propose a bidirectional-guided underwater object detection method, referred to as BG-YOLO (shown in Fig. 1(d)). BG-YOLO consists of image enhancement branch, object detection branch and feature guided module. Specifically, the image enhancement branch and the object detection branch are organized in a parallel way, and the feature guided module connects shallow convolution layer of them, optimizing the network’s training direction through constraining the low-level features of both branches. Our proposed method is significantly different from the parallel method above proposed by Liu et al. [9]. First, in Liu et al.’s method, the image enhancement branch uses a trained network, which is not always conducive to detection tasks. Relatively, in our proposed method, object detection is used to guide the training of the image enhancement branch, optimizing it in a direction conducive to object detection. Therefore, our method has better generalization ability. Besides, Liu et al. used enhanced images to constrain the training of detection branch through consistency loss. However, the enhanced images output from the enhancement branch have essential differences with the shallow feature map from detection branch. By contrast, we constrain the training of detection module through the consistency loss between the low-level features of the two branches.

To summarize, our main contributions are as follows:
\begin{enumerate}
\item{We proposed a object detection framework in underwater degraded scenes, BG-YOLO. Firstly, the detection tasks are used to guide the training of image enhancement network, which makes the enhancement network conducive to detection tasks. Subsequently, the enhancement branch and the detection branch are organized in a parallel way, and the enhancement branch is used to guide the training of detection branch. Finally, during the detection, only the detection branch will be reserved thus no additional computation cost will be introduced.}
\item{We imposed constraints on the corresponding convolutional layer of the enhancement branch and the detection branch, both of which have the same dimension and underlying semantics. This enables the detection branch to learn more feature information and thereby improve the object detection performance of the detection branch.}
\end{enumerate}

Extensive experiments on URPC2019 and URPC2020 demonstrate that, our proposed BG-YOLO significantly improves detection performance compared to the original detection method.

The rest of this article is organized as follows: Section 2 makes a brief review of the underwater image enhancement and the related existing techniques. Section 3 gives a detailed exposition of the proposed method. Section 4 presents sufficient experiments and ablation studies. And the conclusion is illustrated in Section 5.

\section{Related Works}
The research objective of this article is to improve the performance of object detection in complex underwater environment, which mainly involves techniques relevant to underwater image enhancement and object detection. We first review the existing research findings in the field of underwater image enhancement. Then we briefly illustrate the progress of object detection techniques in complex underwater environment. Finally, we focus on the joint optimization of underwater image enhancement and object detection tasks.
\subsection{Underwater Image Enhancement}
Underwater image processing techniques can overcome the problems of image degradation to a large extent. Generally, underwater image enhancement techniques is classified into traditional approaches and deep-learning based approaches.

Underwater image Enhancement includes non-physical model-based methods [11-19] and physical model-based methods [20-22].As the wide application of deep learning in various computer vision tasks, the deep-learning based algorithms have been applied to underwater image enhancement and have achieved remarkable results [23-27]. Li et al. [28] proposed UWCNN, a convolutional neural network model for underwater image enhancement based on underwater image prior, which directly restores clear underwater images. Espinosa et al. [29] combined the discrete wavelet transform (DWT) and proposed a variant of U-Net for underwater image enhancement, in which discrete wavelet transform is utilized in skip connection and is used to achieve de-blurring and colour correction with the channel attention module. Generative adversarial network (GAN) is also widely applied to underwater image enhancement [30-32]. Jiang et al. [33] proposed a domain adaption framework for real world underwater image enhancement, using CycleGAN [34] to transfer the underwater-style images into in-air style and thereby ameliorate the quality of images. Image enhancement methods based on deep-learning can obtain enhanced images with vivid visual effects without estimating prior parameters, but require numerous paired data to train the networks.
\subsection{Underwater Image Detection}
With the development of deep-learning technology, object detection algorithms based on deep-learning are widely used in underwater object detection [35,37,39]. Zeng et al. [36] introduced adversarial occlusion network (AON) into Faster R-CNN, preventing overfitting effectively through adversarial learning, and consequently achieved a more robust detection network for underwater object detection. Cao et al. [38] utilized lightweight MobileNetv2 as the backbone network of SSD algorithm and proposed Faster MSSDLite for underwater detection tasks of live crabs. Liu et al. [40] utilized YOLOv4 as backbone network, using a dual-branch structure of detection branch and tracking branch in parallel way to detect and track marine fish in real time. Yu et al. [41] designed an underwater object detection network U-YOLOv7 based on YOLOv7 to meet the requirements in both speed and precision.

In underwater scenes, the distortion of images is the main factor affecting the performance of object detection. Image enhancement can intuitively improve the visual performance of underwater images. And how to effectively combine the image enhancement algorithms to improve the performance of object detection in underwater low-quality images remains a research objective with significant value.
\subsection{Joint Optimization}
Recent researches integrated the tasks of image enhancement and object detection into one end-to-end framework, optimizing both the image enhancement network and the object detection network jointly when training [1,42-43]. In IA-YOLO [44], a differentiable image processing module DIP is introduced, which uses a small convolutional neural network CNN-PP to predict the parameters of DIP and achieve better detection performance through end-to-end joint learning of CNN-PP and YOLOv3. In DE-YOLO [5], the image enhancement module DENet and the detection module YOLOv3 are organized in cascaded way for joint training. In literature [6], CycleGAN image enhancement module and SSD detection module are integrated into one framework, and the relevant detection information from detector is applied to guide the optimization of the enhancement module towards the direction conducive to detection tasks.

The end-to-end frameworks can enhances the performance of detection tasks, but introduce additional computation cost. DSNet [7] utilizes a dual-subnet structure, in which the recovery subnet and the detection subnet are connected in parallel, sharing a common block. When training, both subnets are trained jointly, but during the detection, only the detection subnet is used. Consequently no additional computation cost will be introduced. In JADSNet [8], a joint attention-guided dual-subnet network is introduced to address the problems in marine object detection through jointly learning image enhancement and object detection tasks. The detection subnet utilizes RetinaNet as backbone to classify and locate the objects, and the image enhancement subnet share the feature extraction layer with the detection subnet. Liu et al. [9] organized the detection branch and the enhancement branch in a parallel way, using enhanced images to guide the low layers of detection branch to learn lost details. This effectively improves the precision of detection while obtaining enhanced output with excellent visual effect, however, not able to ensure that enhancement network achieving significant visual appearance will be most conducive to object detection tasks.

\section{Proposed method}
\subsection{Overview of the Method}
\begin{figure}[ht!]
    \centering
    \includegraphics[width=0.6\linewidth]{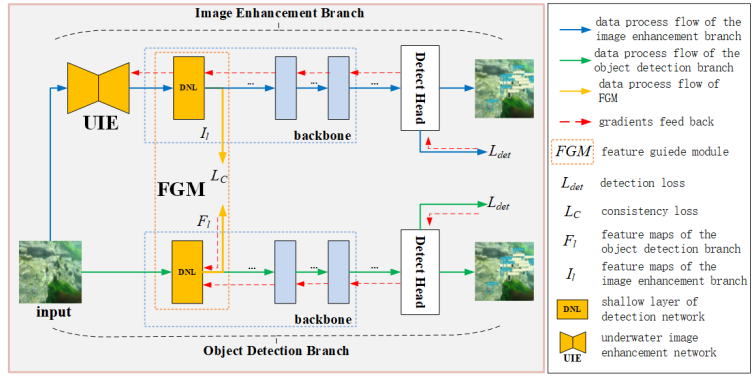}
    \caption{Overview of BG-YOLO framework.}
    \label{fig2:enter-label}
\end{figure}

The research of literature [9] demonstrates that, extracting low-level features is of essential significance for detection in visually degraded scenes. Addressing the image degradation causing difficulties in object detection in complex underwater scenes, we proposed a bidirectional-guided object detection framework, transferring information between image enhancement and object detection. As is shown in Fig. \ref{fig2:enter-label}, the framework consists of three parts: image enhancement branch, object detection branch and feature guided module, and the image enhancement branch and the object detection branch are organized in parallel way. During the training, the features extracted by the enhancement branch is used to guide the detection branch to learn the low-level features and more detailed object information beneficial to the detection tasks and thereby enhance the performance of detection. During the tests, we removed the image enhancement branch and the feature guided module, processing detection with only the trained detection branch. Compared to the method proposed in literature[6], no additional computation cost is introduced to object detection since there’s no cascaded image enhancement network in our proposed framework.

Different from what is proposed in literature [9], considering that individually trained image enhancement networks are not always conducive to detection tasks, we cascade an image enhancement subnet with an object detection subnet as image enhancement branch instead of utilizing pre-trained image enhancement network. When training the image enhancement branch, the detection subnet guides the enhancement subnet towards optimization conducive to object detection tasks. Subsequently, when training the detection branch, the parameters in image enhancement branch are fixed.

Besides, different from the feature guided module in literature [9], ours extracts features from the shallow convolutional layer of the object detection branch and the backbone of the detection subnet in the image enhancement branch respectively. With the constrain of the proposed consistency loss, the low-level features of the object detection branch will tend towards the low-level features of the image enhancement branch gradually, which enables the detection branch to extract more detailed information of objects.

It’s worth mentioning that, different from other joint optimization methods [5-7,9], the objective of our proposed method is achieving better performance of underwater object detection. On the contrary, the visual appearance of the output of the enhancement branch is not taken into consideration.
\subsection{Image Enhancement Branch}
\begin{figure}[ht!]
    \centering
    \includegraphics[width=0.6\linewidth]{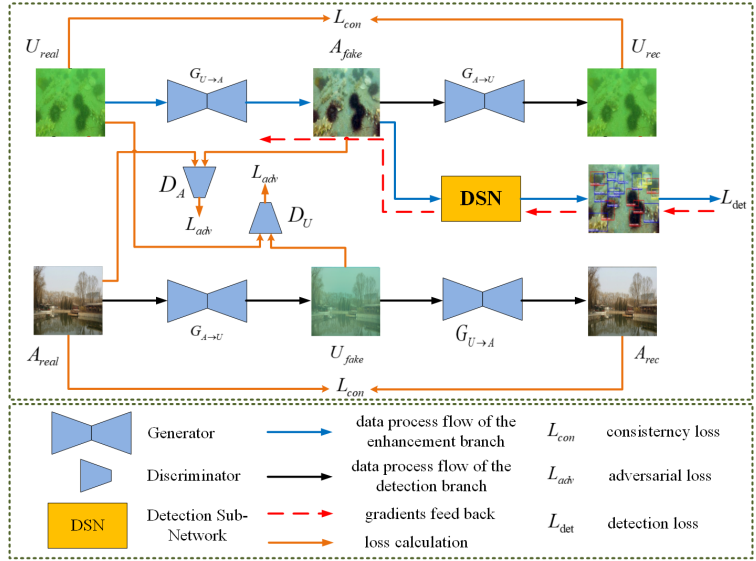}
    \caption{Overview of the image enhancement branch.}
    \label{fig3:enter-label}
\end{figure}

A cascade of the image enhancement subnet and the object detection subnet is utilized as the image enhancement branch, which makes the image enhancement subnet more conducive for ameliorating the performance of object detection subnet. The structure of the image enhancement branch is shown in Fig. \ref{fig3:enter-label}, which consists of an image enhancement subnet and a detection subnet (DSN).

The image enhancement utilizes generators similar to CycleGAN structure. CycleGAN is constructed by two generators: \(G_{U \rightarrow A}\) and \(G_{A \rightarrow U}\), two adversarial discriminators: \(D_{U}\) and \(D_{A}\). The generator \(G_{U \rightarrow A}\) transfers the degraded underwater images \(U_{real}\) from underwater domain to in-air domain, and generate the output images \(A_{fake}\) consequently. The other generator \(G_{A \rightarrow U}\) processes style transfer in the opposite direction, to be precise, transferring the underwater images transferred to in-air domain previously to underwater domain and obtaining the output \(U_{rec}\) consequently. The adversarial discriminator \(D_{A}\) prompts the generator \(G_{U \rightarrow A}\) to transform the original underwater images \(U_{real}\) into output that is difficult to distinguish from real clear in-air images \(A_{real}\). Similarly, prompts the generator \(G_{A \rightarrow U}\) to transform real clear in-air images \(U_{real}\) into output that is difficult to distinguish from underwater images \(U_{real}\).

DSN is designed to describe the internal features and semantic information necessary for detection tasks, emphasizing more features beneficial for detection and transmitting the perception of the detector to the image enhancement subnet, so that the potential output of the image enhancement subnet will be more conducive to detection.

With the intention of making the features extracted by the subsequent feature guided branch more comparable, DSN is implemented using YOLOv5, constitutes the image enhancement branch with the generator \(G_{U \rightarrow A}\) of CycleGAN, as shown by the blue arrows in Fig. \ref{fig2:enter-label} and Fig. \ref{fig3:enter-label}. CycleGAN is trained on dataset containing unpaired underwater and in-air images. The pre-trained generator \(G_{U \rightarrow A}\) is utilized as image enhancement module subsequently. Then the detector is pre-trained with the enhanced data to acquire fundamental knowledge of target classes. Finally, the framework is trained with the original underwater dataset with annotations to achieve appreciable detection performance.

\subsection{Object Detection Branch}
Our proposed framework utilizes YOLOv5 as detection network, for it’s widely used in underwater object detection [45-47]. In our proposed framework, the image enhancement branch is supposed to guide the optimization of the shallow layers of the backbone network of the detection branch. As shown by the orange arrows in Fig. \ref{fig2:enter-label}, multiple low-level features in different scales \(F_{l}, l \in\{1,2,3\}\) are extracted from the backbone network of YOLOv5 in the detection branch. These low-level features  extracted from the detection branch tend to be consistent with the corresponding low-level features of enhancement branch \(I_{l}, l \in\{1,2,3\}\) under the guided of the feature guided branch, which makes the features of objects more salient for detector, and is conducive to ameliorating the performance of object detection.

\subsection{Feature Guided Module}
In the underwater images captured in underwater environment, some features of objects are distorted or even obscured by complex background, and the degradation of the images further worsen it. As a result, these objects may be ignored by the network at the beginning. Consequently, making the shallow features of the detection network tend to be consistent with the image enhancement features conducive to object detection will make the features of the objects more prominent. Propagating these prominent features to the deeper layers of the object detection network can effectively improve the performance of object detection.

\begin{figure}[ht!]
    \centering
    \includegraphics[width=0.6\linewidth]{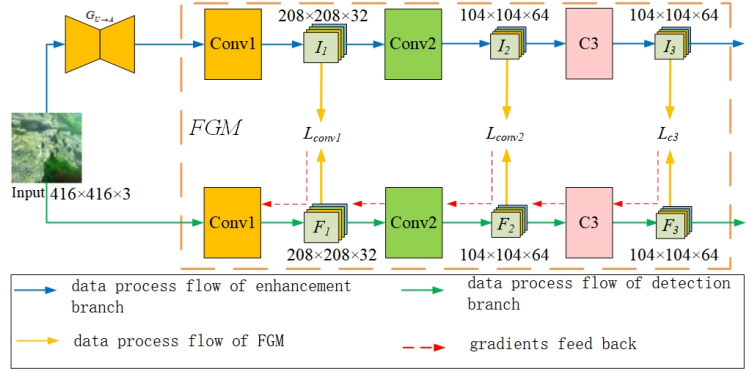}
    \caption{Overview of the feature guided module.}
    \label{fig4:enter-label}
\end{figure}

The feature guided module constrains the low-level features \(F\) extracted by the detection subnet of the object detection branch to converge to \(I\), the low-level features extracted by the image enhancement branch of the detection image. An overview of its structure is shown in Fig. \ref{fig4:enter-label}. The input of the feature guided module originates from two sources: the multi-level shallow feature mappings \(F_{l}, l \in\{1,2,3\}\) extracted by the backbone network of the detection subnet of the object detection branch and the multi-level shallow feature mappings \(I_{l}, l \in\{1,2,3\}\) extracted by the detection subnet of the image enhancement branch. \(F_{l}\) and \(I_{l}\) are extracted from the corresponding layer of object detection subnet, both of which have the same dimension and similar semantic information. Subsequently, minimizing consistency loss is utilized to make the feature mappings, \(F_{l}\) and \(I_{l}\), tend towards consistency. The enhancement subnet is fixed when training to constrain the shallow layers of the detection subnet to optimize in the direction conducive to image enhancement and thereby obtain more detailed information of the objects.

\subsection{Loss Function}
\begin{enumerate}
\item {Image Enhancement Loss}

The image enhancement subnet is designed to generate clear in-air images from the input degraded underwater images. CycleGAN has two generators, \(G_{U \rightarrow A}\) and \(G_{A \rightarrow U}\), and two corresponding adversarial discriminators, \(D_{U}\) and \(D_{A}\). For the mapping \(G_{U \rightarrow A}\) and its corresponding discriminator \(D_{A}\), the adversarial loss can be mathematically expressed as:
\begin{align}
\hspace*{-0.5cm}
L_{GAN}(G_{U \rightarrow A}, D_A, X_U, X_A) = & \nonumber \\
& \hspace*{-2.5cm} {E}_{x_a \sim p_{data}(x_a)}[\log D_A(x_a)] \nonumber \\
& \hspace*{-2.5cm} +{E}_{x_u \sim p_{data}(x_u)}[\log(1 - D_A(G_{U \rightarrow A}(x_u))]
\end{align}

where mapping \(G_{U \rightarrow A}\) is supposed to generate images  \(G_{U \rightarrow A}(X_{u})\) similar to images in in-air domain \(X_{A}\) using images in underwater domain \(X_{U}\), and the discriminator \(D_{A}\), is expected to distinguish between the generated in-air domain images \(G_{U \rightarrow A}(X_{u})\) and the real in-air domain images \(X_{a}\). Similarly, for the mapping \(G_{A \rightarrow U}\) and its corresponding discriminator \(D_{U}\), the adversarial loss can be mathematically expressed as:
\begin{align}
\hspace*{-0.5cm}
L_{GAN}(G_{A \rightarrow U}, D_U, X_A, X_U) = & \nonumber \\ 
& \hspace*{-2.5cm}{E}_{x_u \sim p_{data}(x_u)}[\log D_U(x_u)]\nonumber \\
& \hspace*{-2.5cm} +{E}_{x_a \sim p_{data}(x_a)}[\log(1 - D_U(G_{A \rightarrow U}(x_a)))]
\end{align}

The cycle consistency loss prevents the learnt mappings \(G_{U \rightarrow A}\) and \(G_{A \rightarrow U}\) from being contradictory. For each image \(X_{u}\) from domain \(X_{U}\), the forward cycle of image transfer is supposed to transform \(X_{u}\) back into the original image \(X_{u}\), in other words, . Similarly, for each image \(X_{a}\) from the domain \(X_{A}\), there exists . The consistency loss [33] can be expressed as:
\begin{align}
\hspace*{-0.5cm}
L_{con} \left( G_{U \to A}, G_{A \to U} \right) = & \nonumber \\ 
& \hspace*{-2.5cm}{E}_{x_u \sim {P}_{data}(x_u)} \left[ \left\| G_{A \to U} \left( G_{U \to A} \left( x_u \right) \right) - x_u \right\|_1 \right] \nonumber \\
& \hspace*{-2.5cm} +{E}_{x_a \sim {P}_{data}(x_a)} \left[ \left\| G_{U \to A} \left( G_{A \to U} \left( x_a \right) \right) - x_a \right\|_1 \right]
\end{align}

The cycle consistency loss [48] is introduced to maintain the composition of original image through extracting features in both high and low levels using VGG-16. The cycle consistency loss can be mathematically expressed as:
\begin{align}
\hspace*{-0.5cm}
L_{cp}(G_{U \rightarrow A},\, G_{A \rightarrow U}) = & \nonumber \\
& \hspace*{-2.5cm} \| \phi(x_u) - \phi(G_{A \rightarrow U}(G_{U \rightarrow A}(x_u))) \|_2^2 \nonumber \\
& \hspace*{-2.5cm} + \| \phi(x_a) - \phi(G_{U \rightarrow A}(G_{A \rightarrow U}(x_a))) \|_2^2
\end{align}

The total loss of image enhancement \(L_{UIE}\) can be expressed as:
\begin{align}
\hspace*{-0.5cm}
L_{UIE}\left(G_{U \to A}, G_{A \to U}, D_A, D_U\right) = & \nonumber \\
& \hspace*{-2.5cm} L_{GAN}\left(G_{U \to A}, D_A, X_U, X_A\right) \nonumber \\
& \hspace*{-2.5cm} + L_{GAN}\left(G_{A \to U}, D_U, X_A, X_U\right) \nonumber \\
& \hspace*{-2.5cm} + \lambda_1 L_{con}\left(G_{U \to A}, G_{A \to U}\right) \nonumber \\
& \hspace*{-2.5cm} + \lambda_2 L_{cp}\left(G_{U \to A}, G_{A \to U}\right)
\end{align}

where \(\lambda_1\), \(\lambda_2\) control the relative importance of the two losses.

\item{Object Detection Loss}

Both the object detection module of the image enhancement branch and the object detection branch utilize the same structure of YOLOv5s and consequently use the loss functions same as original YOLOv5s, as is shown below:
\begin{align}
\hspace*{-0.5cm}L_{det} & = a \cdot \text{LOSS}_{obj} + b \cdot \text{LOSS}_{loc} + c \cdot \text{LOSS}_{cls}
\end{align}
where \({LOSS}_{obj}\) is the confidence loss, a function on probability that an object exists, more specifically, whether an object exists inside the predicted bounding box; \({LOSS}_{obj}\) is the classification loss, a binary cross-entropy function on the predicted probability that the object in a bounding box belongs to a class and the ground truth; and \({LOSS}_{loc}\) is a function measuring the difference between the predicted bounding boxes and the ground truth. And a, b, c are the weight factors respectively.

\item {Consistency Loss}

To measure the consistency of low-level features of the object detection branch and the enhancement branch, we utilize mean square error (MSE) function as consistency loss. MSE, convex and differentiable, is widely used in metrics in regression, pattern recognition, signal and image processing, etc. We use MSE to measure the difference between the feature maps in pixel-level and attempt to minimize it. With consistency loss, the detection network is able to understand the subtle features of the distribution of objects.

The feature-guided consistency loss is expressed as:
\begin{align}
L_{con} = \frac{1}{h \cdot w} \sum_{i = 0}^{h-1} \sum_{j = 0}^{w-1} \left[F_l(i,j) - I_l(i,j)\right]^2
\end{align}

where \(h\), \(w\) are the height and width of the feature map, and \(F_l(i,j)\), \(I_l(i,j)\) represent the features from the l-th level of the object detection subnet and the image enhancement subnet respectively. The full guided loss is mathematically expressed as:
\begin{align}L_{FGM} & = \mu_1 L_{con1} + \mu_2 L_{con2} + \mu_3 L_{c3}\end{align}

where \(\mu _{l}, l \in\{1,2,3\}\) is used to balance the consistency loss between the different feature layers \((F_l, I_l), l \in \{1,2,3\}\).

\item{Total Loss Function}

The image enhancement branch and the object detection branch are trained separately. When training the image enhancement branch, first separately train the image enhancement subnet and the object detection subnet, thus the loss functions are respectively the image enhancement loss and the object detection loss. Subsequently, only the object detection loss is used when jointly optimizing the image enhancement subnet.

When training the object detection branch, the parameters of the image enhancement branch are fixed, and the total loss of the object detection branch  is defined as sum of the detection loss \(L_{det}\) and the consistency loss \(L_{FGM}\), as is shown below:
\begin{align}L & = \eta_1 L_{det} + \eta_2 L_{FGM}\end{align}
where \(\eta_1\) and \(\eta_2\) are balance factors.
\end{enumerate}

\section{Experiments}
\subsection{Datasets}
We adopt publicly accessible datasets, Underwater Robot Picking Contest 2019 (URPC2019) and Underwater Robot Picking Contest 2020 (URPC2020), to evaluate the performance of our proposed object detection framework. The URPC datasets are widely used to evaluate the performance of object detection method in underwater scenes, which can be downloaded from http://www.cnurpc.org.

The dataset on URPC2019 contains 3765 images in train set and 942 in test set, covering 5 classes of underwater targets: echinus, starfish, holothurian, scallop and waterweeds. The images in the dataset present adverse characteristics including colour deviation, blurriness, low contrast, clustered objects and occlusion.

The dataset on URPC2020 includes a train set consisting of 4200 randomly chosen images and a test set consisting of 800 randomly chosen images. The URPC2020 dataset includes 4 different kinds of underwater targets: holothurian, echinus, scallop and starfish.
\subsection{Implementation Details}
We implemented our framework using Python-3.8.10, Torch-1.10.0 with IntelXeon(R) Platinum 8255C CPU@2.50GHz, 43GB memory and Nvidia GeForce RTX3090.

When training BG-YOLO, first apply joint optimization on the image enhancement branch, and then use the pre-trained model to guide the training of the object detection branch. The image enhancement subnet is publicly released CycleGAN network model, and the detection subnet is publicly released YOLOv5 model.
\begin{enumerate}
\item{Training of the Image Enhancement Branch}

To train the image enhancement branch, separate training of the image enhancement subnet and the object detection subnet should be first conducted. And subsequently is the joint optimization of the two.

When training the image enhancement branch, the training details of reference [33] and [6] are partly referenced. The images for training are from the dataset UIEB [49] and the dataset EUVP [50]. UIEB contains 890 pairs of images, with each pair containing one clear image and its corresponding blurred one. And EUVP contains 11435 degraded/clear images in various background. We randomly chose 400 unpaired degraded/clear image from UIEB and 600 from EUVP as training set. The images are then scaled into 416×416. During the experiments, we found that when we choose the clear in-air image as target domain image, the transferred image suffers from obvious artifacts, distortion and checkerboard effects, which severely affect the performance of detector. We finally chose the clear underwater images as target domain. When training the image enhancement subnet, Adam optimizer is used. We trained the subnet for 50 epochs, with batch size = 2, momentum \(\beta_1=0.5\), learning rate lr = 1e-4, and the weights \(\lambda _1=5e-5\), \(\lambda _2=1\).

When training the detection subnet of the image enhancement branch, we chose the optimal configurations according to the test results referring the training details of reference [6]. The datasets used are URPC2019 and URPC2020. We first use the pre-trained enhancement subnet mentioned above to enhance the images from URPC datasets and then the enhanced images will serve as the input of the detection subnet. The optimizer utilized is SGD. We trained the subnet based on publicly released pre-trained model yolov5s.pt for 300 epochs, with batch size = 16, momentum = 0.937, learning rate lr = 1e-2, lrf = 1e-2, weight decay = 5e-4.

When processing joint optimization on the image enhancement branch, we chose the optimal configurations according to the test results referring the training details of reference [6]. The datasets used are URPC2019 and URPC2020. When training, the pre-trained enhancement subnet model and the pre-trained detection subnet model mentioned above is first loaded. The optimizer used is SGD. We trained the branch for 300 epochs with batch size = 16, momentum = 0.937, learning rate lr = 1e-3, lrf = 1e-3 and weight decay = 5e-4.

\item{Training of the Image Enhancement Branch}

The object detection branch consists of only one detection subnet. When training, the pre-trained model of the image enhancement branch obtained by joint optimization is loaded. The parameters of the enhancement branch are fixed and optimization is only applied to the object detection branch. The datasets used are URPC2019 and URPC2020. And the optimizer used is SGD. We trained the branch for 300 epochs with batch size = 16, momentum = 0.937, learning rate lr = 1e-2, lrf = 1e-2 and weight decay = 5e-4. The object detection branch adopts publicly released pre-trained model yolov5s.pt.
\item{Comparison}

To evaluate the performance of our proposed method, we reproduced the corresponding proposed algorithms according to the details in references [6] and [9] for comparison. However, we utilized YOLOv5s as object detection network in all these algorithms to control variables and facilitate comparison, which will not affects the final conclusion.
\end{enumerate}
\subsection{Evaluation Indices}
To comprehensively and objectively evaluate the performance of our proposed method, we use mean of average precision (mAP), recall, precision, F1-score, PR curve and detection speed (FPS). When testing the detection speed, the batch size is set as 1.

\subsection{Visualized Comparison}
\begin{figure}[htbp]
    \centering
    \includegraphics[width=0.7\linewidth]{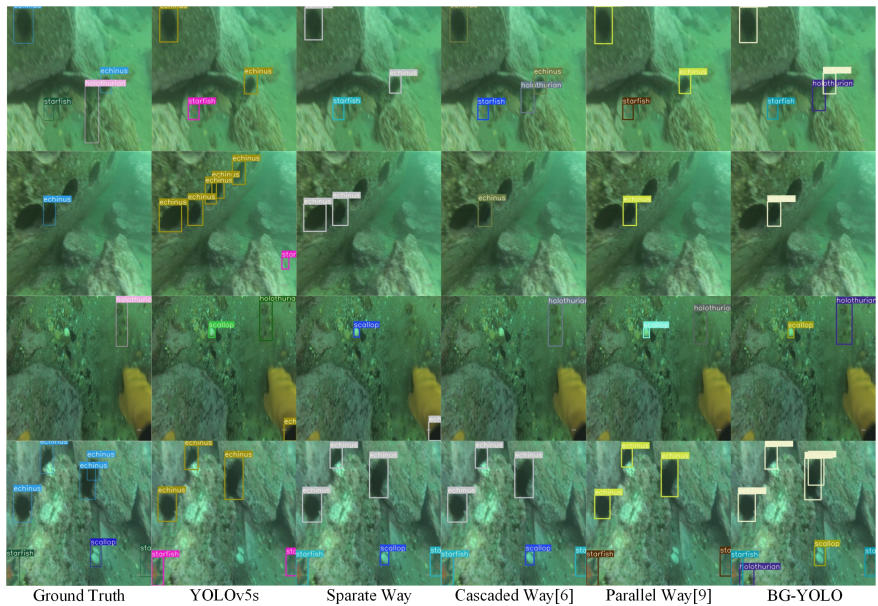}
    \caption{Visualized detection results of different methods on URPC2019 dataset.}
    \label{fig5:enter-label}
\end{figure}

In this section, we first compare the visualized detection results of the original YOLOv5s, the separate way, the cascaded way, the parallel way and the proposed BG-YOLO on URPC2019 dataset. The visualized results of Fig. \ref{fig5:enter-label} show that, our proposed method can better detect clustered and occluded objects in degraded underwater images.

\begin{figure}[htbp]
    \centering
    \includegraphics[width=0.7\linewidth]{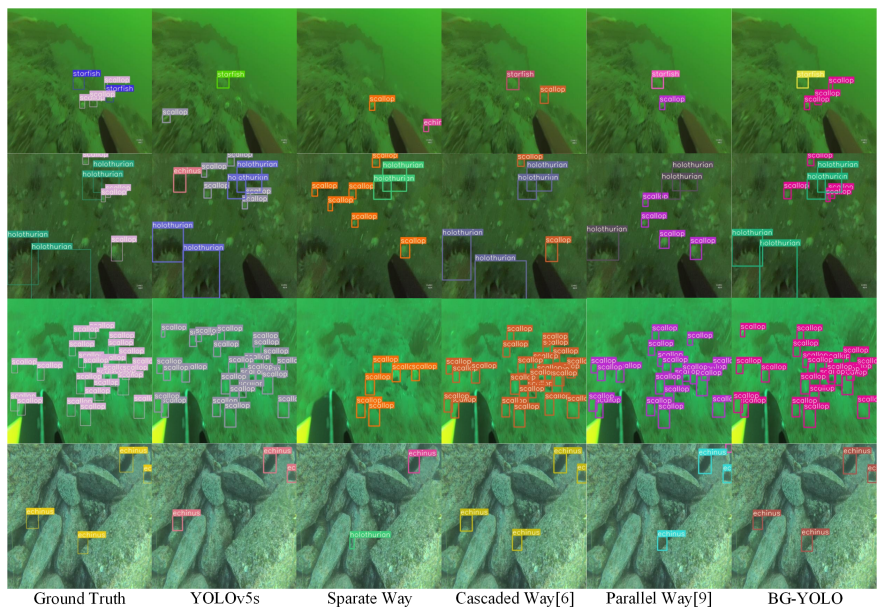}
    \caption{Visualized detection results of different methods on URPC2020 dataset.}
    \label{fig6:enter-label}
\end{figure}

Fig. \ref{fig6:enter-label} exhibits the visualized results of 4 images sampled from URPC2020 dataset. It can be seen that BG-YOLO shows better detection performance in severely degraded underwater scenes, regardless of whether the objects are small, clustered, or occluded.

\subsection{Quantitative Comparison}
\begin{enumerate}
\item{Results on URPC2019 Dataset}

We utilized third-party publicly released YOLOv5s as basic network model. We conducted tests on URPC2019 dataset, comparing the detection performance of the original YOLOv5s, the separate way, the cascaded way, the parallel way and BG-YOLO. When testing, we first resized the images to 416×416, and the division of dataset was the same as the original dataset, with 3765 images in training set and 942 in the validation set (test set). The cascaded way and the parallel way are reproductions of the methods proposed in references [6] and [9], respectively. The test results are shown in Table \ref{tab1:my_label}.
\begin{table}[ht!]
    \centering
    \caption{Test Results on URPC2019}
    \label{tab1:my_label}
    \small 
    \begin{tabular}{ccccccc}
        \toprule 
        method & mAP@0.5(\%) & mAP@0.5-0.95(\%) & Recall(\%) & F1-score(\%) & Precision(\%) & FPS(fps) \\
        \midrule 
        YOLOv5s & 75.3 & 42.0 & 72.6 & 74.0 & 74.8 & 130 \\
        Separate Way & 74.7 & 39.6 & 68.6 & 75.0 & 83.4 & 130 \\
        Cascaded Way [6] & 76.9 & 42.8 & 72.7 & 76.0 & 81.0 & 42 \\
        Parallel Way [9] & 73.1 & 41.7 & 66.3 & 72.0 & 78.7 & 130 \\
        BG-YOLO & 78.4 & 44.7 & 70.1 & 77.0 & 88.3 & 130 \\
        \bottomrule 
    \end{tabular}
\end{table}

From the Table \ref{tab1:my_label}, it can be seen that the mAP@0.5 of our proposed BG-YOLO is 78.4\%, and the mAP@0.5-0.95 is 44.7\%, which is an improvement 0f 3.1\% and 2.7\% respectively comparing to the baseline YOLOv5s. Compared to the separate way, the cascaded way and the parallel way, BG-YOLO achieves an improvement of 3.7\% and 5.1\%, 0.6\% and 4.4\%, 1.5\% and 2.6\%, respectively. Besides, the F1-score of BG-YOLO is also the best among the methods to compare mentioned above, which demonstrates that BG-YOLO achieves an optimal balance between detection precision and recall. Through the analysis of the visualized detection results (shown Figure 5), the original YOLOv5s achieves higher recall rate than BG-YOLO, however, its false drop rate in detection is also higher. The separate way and the cascaded way achieve higher detection precision than BG-YOLO, but their miss rate in detection is higher. Additionally, the inference speed of BG-YOLO is also the fastest among these methods, which reaches 130fps, while the speed of the cascaded way is only 42fps because of computational cost brought by combining UIE and UOD into one framework.

From Fig. \ref{fig7:enter-label}, it can be seen that the area enclosed by the PR curve of BG-YOLO (purple) and the coordinate axes is the largest, indicating that BG-YOLO achieves the best performance among these methods.
\begin{figure}[ht!]
    \centering
    \includegraphics[width=0.5\linewidth]{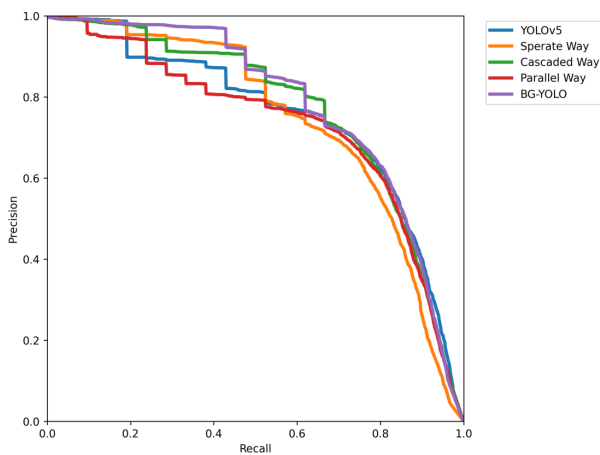}
    \caption{PR curve of the test results on URPC2019.}
    \label{fig7:enter-label}
\end{figure}

\item{Results on URPC2020 Dataset}

We further conducted tests on URPC2020 in the same way as the previous section. The test results are shown in Table \ref{tab2:my_label}.

\begin{table}[ht!]
    \centering
    \caption{Test Results on URPC2020}
    \label{tab2:my_label}
    \small
    \begin{tabular}{ccccccc}
        \toprule 
        method & mAP@0.5(\%) & mAP@0.5-0.95(\%) & Recall(\%) & F1-score(\%) & Precision(\%) & FPS(fps) \\
        \midrule 
        YOLOv5s & 79.5 & 44.9 & 73.8 & 78.0 & 83.7 & 132 \\
        Separate Way & 74.8 & 40.3 & 69.1 & 75.0 & 81.1 & 132 \\
        Cascaded Way [6] & 80.4 & 44.8 & 75.7 & 79.0 & 82.0 & 42 \\
        Parallel Way [9] & 75.9 & 38.4 & 68.1 & 73.0 & 79.7 & 132 \\
        BG-YOLO & 80.2 & 44.6 & 75.2 & 79.0 & 82.7 & 132 \\
        \bottomrule 
    \end{tabular}
\end{table}

As is shown in Table \ref{tab2:my_label}, the mAP@0.5 of our proposed BG-YOLO is 80.2\%, which is 0.7\% higher than the baseline, and 5.4\%, 4.3\% higher than the separate way and the parallel way respectively. Though slightly lower than the cascaded way (0.2\%) in mAP@0.5, BG-YOLO is three times as fast in terms of speed. The PR curves shown in Fig. 8 also demonstrate the performance of BG-YOLO in comparison to other methods.
\begin{figure}[ht!]
    \centering
    \includegraphics[width=0.5\linewidth]{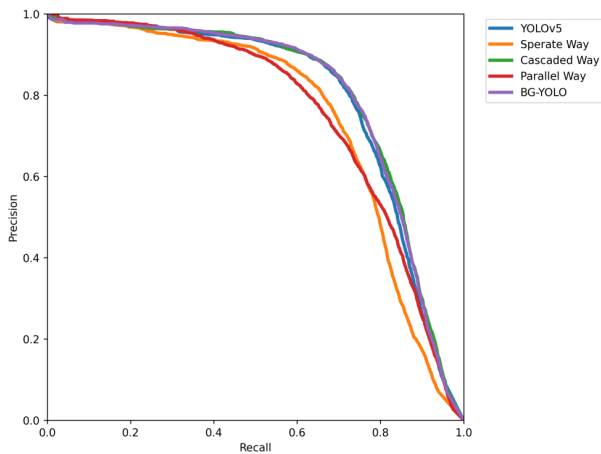}
    \caption{PR curve of the test results on URPC2020.}
    \label{fig8:enter-label}
\end{figure}

It’s worth mentioning that the performance of the reproduced parallel approach referring literatrue [9] is even worse than original YOLOv5s. This is because we didn’t adopt particular enhancement algorithm when reproducing it, which also demonstrates that the performance of the litreature [9] depends on the enhancement algorithm chosen.

\item{Effects of Feature guided Layer}

To evaluate the influence of different shallow convolutional layers \(f_{1}\) and \(I_{1}\), we tested the difference in guiding the training of BG-YOLO by extracting features \((F_l, I_l), l \in \{1,2,3\}\) from the first convolutional layer conv1, the second convolutional layer conv2, and the C3 module of the enhancement branch and the corresponding detection network in the detection branch on the URPC2019 dataset. The results are presented in Table \ref{tab3:my_label}. In this part of experiments, the weight factors of detection loss \(L_{det}\) and consistency loss \(L_{FGM}\) are \(\lambda _1=\lambda _2=1.0\), and the weight factors of \(L_{con1}\), \(L_{con2}\) and \(L_{C3}\) are \(\mu _1=\mu _2=\mu _3=1.0\).

\begin{table}[ht!]
    \centering
    \caption{Contribution of Feature guided Layers to Detection Precision, ``\(\sqrt{}\)" Indicates that the operation has been performed}
    \label{tab3:my_label}
    \small
    \begin{tabular}{ccccc}
        \toprule
        conv1 & conv2 & C3 & mAP@0.5 & mAP@0.5-0.95 \\
        \midrule
        & & & 75.3 & 42.0 \\
        \(\sqrt{}\)& & & 78.0 & 43.8 \\
        & \(\sqrt{}\)& & 77.7 & 44.2 \\
        & & \(\sqrt{}\)& 77.0 & 43.8 \\
        \(\sqrt{}\)& \(\sqrt{}\)& & 76.0 & 43.5 \\
        \(\sqrt{}\)& & \(\sqrt{}\)& 74.8 & 43.2 \\
        & \(\sqrt{}\)& \(\sqrt{}\)& 76.6 & 43.5 \\
        \(\sqrt{}\)& \(\sqrt{}\)& \(\sqrt{}\)& 76.6 & 43.6 \\
        \bottomrule
    \end{tabular}
\end{table}

As shown in Table \ref{tab3:my_label}, each combination of the feature guided layers presented above contributes to improving the detection performance, which indicates that guiding the low-level features of the detection network towards image enhancement is conducive to detection tasks. When using only the features of the first convolutional layer for guided, the best performance of detection is achieved. In degraded underwater scenes, the main reason leading to the decline in object detection performance is that the degraded underwater images lack detailed features conducive to object detection. Besides, deeper convolutional layers focus more on semantic information, which does not significantly contribute to improving the performance of object detection performance.

\item{Effects of Different \(\eta \)}

According to formula (9), the values of \(\eta _1\) and \(\eta _2\) balance the effects of the object detection branch and the enhancement branch. We fixed \(\eta _1\) and tested the contribution of guiding of the enhancement branch with different values of \(\eta _2\). Considering the results of the previous section, only the shallowest convolutional layer was used for feature guided. And the dataset used in this part of experiments is URPC2019. The results are presented in Table \ref{tab4:my_label}.

\begin{table}[ht!]
    \centering
    \caption{Detection Results with Different \(\eta _2\)}
    \label{tab4:my_label}
    \small
    \begin{tabular}{ccc}
        \toprule
        \(\eta _2\) & mAP@0.5 & mAP@0.5-0.95 \\
        \midrule
        1.0 & 78.0 & 43.8 \\
        0.5 & 77.4 & 44.6 \\
        0.1 & 77.6 & 43.6 \\
        0.05 & 78.4 & 44.7 \\
        \bottomrule
    \end{tabular}
\end{table}

The results presented in Table \ref{tab4:my_label} indicates that, it’s favourable for detection tasks when \(\eta _2\) varies from 0.01 to 1.0, and when \(\eta _2=0.05\), the model reaches highest mAP@0.5, 78.4\%.

\end{enumerate}

\section{Conclusion}
The degradation of images is a significant challenge for underwater object detection tasks. Common approaches focus on ameliorate the visual performance of underwater images through image enhancement methods. However, images with excellent visual performance may not favourable for improving the performance of detection. Although the end-to-end joint optimization adapts the enhancement models to the detection tasks, it not only hinders the model from exerting the abilities of extracting features but also increases the complexity of the network and consequently decrease the inference speed. The parallel organization approach utilizes the enhancement images to guide the learning of the shallow layers of network, but its performance heavily relies on the images used for guided.

Addressing the above issues, we propose a bidirectional-guided object detection method, which combines the advantages of both cascaded organization and parallel organization of image enhancement and object detection network. More specifically, we organize the enhancement branch and the detection branch, and the enhancement branch is a combination of the enhancement subnet and the detection subnet. The detection branch contains only one detection subnet. The feature guided module guides the low-level features to optimize towards the corresponding low-level features of the enhancement branch, in order to make the detection branch sensitive to quality of images and object detection. During the inference, the enhancement branch and the feature guided module are removed, and only the detection branch is used. Extensive experiments demonstrate that, our proposed framework can significantly improve the precision of underwater object detection without introducing additional computation cost, which demonstrates the effectiveness of our proposed method.

\section*{Funding}
Natural Science Foundation of Hainan Province(623RC449); Key Research and Development Project of Hainan Province(ZDYF2024SHFZ089).

\section*{Disclosures}
The authors declare no conflicts of interest.

\section*{Data Availability}
Data underlying the results presented in this paper are not publicly available at this time but maybe obtained from the authors upon reasonable request.

\bibliographystyle{unsrt}  
\bibliography{references}  

Chen L, Jiang Z, Tong L, et al. Perceptual underwater image enhancement with deep learning and physical priors[J]. IEEE Transactions on Circuits and Systems for Video Technology, 2020, 31(8): 3078-3092.

Xu S, Zhang M, Song W, et al. A systematic review and analysis of deep learning-based underwater object detection[J]. Neurocomputing, 2023, 527: 204-232.

Peng W Y, Peng Y T, Lien W C, et al. Unveiling of how image restoration contributes to underwater object detection[C]. 2021 IEEE International Conference on Consumer Electronics-Taiwan (ICCE-TW), Penghu, Taiwan, China, 2021: 1-2.

Liu H, Song P, Ding R. Towards domain generalization in underwater object detection[C]. 2020 IEEE International Conference on Image Processing (ICIP), Abu Dhabi, United Arab Emirates, 2020: 1971-1975.

Qin Q, Chang K, Huang M, et al. DENet: Detection-driven Enhancement Network for Object Detection Under Adverse Weather Conditions[C]. Proceedings of the Asian Conference on Computer Vision. Macau SAR, China, 2022: 2813-2829.

Liu R, Jiang Z, Yang S, et al. Twin adversarial contrastive learning for underwater image enhancement and beyond[J]. IEEE Transactions on Image Processing, 2022, 31: 4922-4936.

Huang S C, Le T H, Jaw D W. DSNet: Joint semantic learning for object detection in inclement weather conditions[J]. IEEE transactions on pattern analysis and machine intelligence, 2020, 43(8): 2623-2633.

Cheng N, Xie H, Zhu X, et al. Joint image enhancement learning for marine object detection in natural scene[J]. Engineering Applications of Artificial Intelligence, 2023, 120: 105905.

Liu H, Jin F, Zeng H, et al. Image Enhancement Guided Object Detection in Visually Degraded Scenes[J]. IEEE Transactions on Neural Networks and Learning Systems, 2023.

Vasamsetti S, Mittal N, Neelapu B C, et al. Wavelet based perspective on variational enhancement technique for underwater imagery[J]. Ocean Engineering, 2017, 141: 88-100.

R. Hummel, Image enhancement by histogram transformation, Comput.Graphics Image Process. 6 (1977) 184–195.

Ketcham D J, Lowe R W, Weber J W. Image enhancement techniques for cockpit displays[J]. Hughes Aircraft Co Culver City Ca Display Systems Lab, 1974.

Pizer S M, Amburn E P, Austin J D, et al. Adaptive histogram equalization and its variations[J]. Computer vision, graphics, and image processing, 1987, 39(3): 355-368.

E.H. Land, J.J. McCann, Lightness and Retinex theory, J. Opt. Soc. Am. 61 (1971) 1–11.

K. Iqbal, M. Odetayo, A. James, R.A. Salam, A.Z.H. Talib, Enhancing the low quality images using Unsupervised Colour Correction Method, in: 2010 IEEE International Conference on Systems, Man and Cybernetics, 2010: pp. 1703–1709.

D. Huang, Y. Wang, W. Song, J. Sequeira, S. Mavromatis, Shallow-water image enhancement using relative global histogram stretching based on adaptive parameter acquisition, in: K. Schoeffmann, T.H. Chalidabhongse, C.W. Ngo, S. Aramvith, N.E. O’Connor, Y.-S. Ho, M. Gabbouj, A. Elgammal (Eds.), MultiMedia Modeling, Springer International Publishing, Cham, 2018, pp. 453–465.

S. Zhang, T. Wang, J. Dong, H. Yu, Underwater image enhancement via extended multi-scale Retinex, Neurocomputing 245 (2017) 1–9.

Liu K, Li X. De-hazing and enhancement method for underwater and low-light images[J]. Multimedia Tools and Applications, 2021, 80: 19421-19439.

Zhang W, Dong L, Xu W. Retinex-inspired color correction and detail preserved fusion for underwater image enhancement[J]. Computers and Electronics in Agriculture, 2022, 192: 106585.

He K, Sun J, Tang X. Single image haze removal using dark channel prior[J]. IEEE transactions on pattern analysis and machine intelligence, 2010, 33(12): 2341-2353.

Galdran A, Pardo D, Picón A, et al. Automatic red-channel underwater image restoration[J]. Journal of Visual Communication and Image Representation, 2015, 26: 132-145.

P. Drews Jr, E. do Nascimento, F. Moraes, S. Botelho, M. Campos, Transmission Estimation in Underwater Single Images, in: 2013 IEEE International Conference on Computer Vision Workshops, 2013: pp. 825–830.

Guo P, Zeng D, Tian Y, et al. Multi-scale enhancement fusion for underwater sea cucumber images based on human visual system modelling[J]. Computers and Electronics in Agriculture, 2020, 175: 105608.

Gangisetty S, Rai R R. FloodNet: Underwater image restoration based on residual dense learning[J]. Signal Processing: Image Communication, 2022, 104: 116647.

Xu S, Zhang J, Qin X, et al. Deep retinex decomposition network for underwater image enhancement[J]. Computers and Electrical Engineering, 2022, 100: 107822.

Wu S, Luo T, Jiang G, et al. A two-stage underwater enhancement network based on structure decomposition and characteristics of underwater imaging[J]. IEEE Journal of Oceanic Engineering, 2021, 46(4): 1213-1227.

Xue X, Hao Z, Ma L, et al. Joint luminance and chrominance learning for underwater image enhancement[J]. IEEE Signal Processing Letters, 2021, 28: 818-822.

Li C, Anwar S, Porikli F. Underwater scene prior inspired deep underwater image and video enhancement[J]. Pattern Recognition, 2020, 98: 107038.

Espinosa A R, McIntosh D, Albu A B. An Efficient Approach for Underwater Image Improvement: Deblurring, Dehazing, and Color Correction[C]. Proceedings of the IEEE/CVF Winter Conference on Applications of Computer Vision.Waikoloa, HI, USA , 2023: 206-215.

Zhang H, Sun L, Wu L, et al. DuGAN: An effective framework for underwater image enhancement[J]. IET Image Processing, 2021, 15(9): 2010-2019.

Sun B, Mei Y, Yan N, et al. UMGAN: Underwater Image Enhancement Network for Unpaired Image-to-Image Translation[J]. Journal of Marine Science and Engineering, 2023, 11(2): 447.

Li J, Skinner K A, Eustice R M, et al. WaterGAN: Unsupervised generative network to enable real-time color correction of monocular underwater images[J]. IEEE Robotics and Automation letters, 2017, 3(1): 387-394.

Jiang Q, Zhang Y, Bao F, et al. Two-step domain adaptation for underwater image enhancement[J]. Pattern Recognition, 2022, 122: 108324.

Zhu J Y, Park T, Isola P, et al. Unpaired image-to-image translation using cycle-consistent adversarial networks[C]//Proceedings of the IEEE international conference on computer vision. 2017: 2223-2232.

Huang H, Zhou H, Yang X, et al. Faster R-CNN for marine organisms detection and recognition using data augmentation[J]. Neurocomputing, 2019, 337: 372-384.

Zeng L, Sun B, Zhu D. Underwater target detection based on Faster R-CNN and adversarial occlusion network[J]. Engineering Applications of Artificial Intelligence, 2021, 100: 104190.

Hu K, Lu F, Lu M, et al. A marine object detection algorithm based on SSD and feature enhancement[J]. Complexity, 2020, 2020: 1-14.

Cao S, Zhao D, Liu X, et al. Real-time robust detector for underwater live crabs based on deep learning[J]. Computers and Electronics in Agriculture, 2020, 172: 105339.

Wang L, Ye X, Xing H, et al. Yolo nano underwater: A fast and compact object detector for embedded device[C]. Global Oceans 2020: Singapore–US Gulf Coast. IEEE,Singapore, 2020: 1-4.

Liu T, Li P, Liu H, et al. Multi-class fish stock statistics technology based on object classification and tracking algorithm[J]. Ecological Informatics, 2021, 63: 101240.

Yu G, Cai R, Su J, et al. U-YOLOv7: A network for underwater organism detection[J]. Ecological Informatics, 2023, 75: 102108.

Zhang X, Fang X, Pan M, et al. A marine organism detection framework based on the joint optimization of image enhancement and object detection[J]. Sensors, 2021, 21(21): 7205.

Yeh C H, Lin C H, Kang L W, et al. Lightweight deep neural network for joint learning of underwater object detection and color conversion[J]. IEEE Transactions on Neural Networks and Learning Systems, 2021.

Liu W, Ren G, Yu R, et al. Image-adaptive YOLO for object detection in adverse weather conditions[C]. Proceedings of the AAAI Conference on Artificial Intelligence, Vancouver, Canada, 2022, 36(2): 1792-1800.

Lei F, Tang F, Li S. Underwater target detection algorithm based on improved YOLOv5[J]. Journal of Marine Science and Engineering, 2022, 10(3): 310.

Wang H, Zhang S, Zhao S, et al. Real-time detection and tracking of fish abnormal behavior based on improved YOLOV5 and SiamRPN++[J]. Computers and Electronics in Agriculture, 2022, 192: 106512.

Hua X, Cui X, Xu X, et al. Underwater object detection algorithm based on feature enhancement and progressive dynamic aggregation strategy[J]. Pattern Recognition, 2023, 139: 109511.

Engin D, Genç A, Kemal Ekenel H. Cycle-dehaze: Enhanced cyclegan for single image dehazing[C]. Proceedings of the IEEE conference on computer vision and pattern recognition workshops, Salt Lake City, UT, USA, 2018: 825-833.

Ancuti C, Ancuti C O, Haber T, et al. Enhancing underwater images and videos by fusion[C]. 2012 IEEE conference on computer vision and pattern recognition, Providence, RI, USA, 2012: 81-88.

Islam M J, Xia Y, Sattar J. Fast underwater image enhancement for improved visual perception[J]. IEEE Robotics and Automation Letters, 2020, 5(2): 3227-3234.

\end{document}